\documentclass[journal]{IEEEtran}

\ifCLASSOPTIONcompsoc
  \usepackage[nocompress]{cite}
\else
  \usepackage{cite}
\fi
\ifCLASSINFOpdf
\else
 \fi
\usepackage{times}
\usepackage{epsfig}
\usepackage{graphicx}
\usepackage{amsmath}
\usepackage{amssymb}

\begin{document}
\title{Learning Deep Features for One-Class Classification}

\author{Pramuditha~Perera,~\IEEEmembership{Student Member,~IEEE,}
        and~Vishal~M.~Patel,~\IEEEmembership{Senior Member~,~IEEE}
\IEEEcompsocitemizethanks{\IEEEcompsocthanksitem P. Perara and V. M. Patel are with the Department of Electrical and Computer Engineering at Johns Hopkins University, MD 21218.\protect\\
E-mail: \{pperera3, vpatel36\}@jhu.edu
}
\thanks{ }}

	\maketitle

\begin{abstract}
	We present a novel deep-learning based approach for \textit{one-class transfer learning} in which labeled data from an unrelated task is used for feature learning in one-class classification.  The proposed method operates on top of a Convolutional Neural Network (CNN) of choice and produces descriptive features while maintaining a low intra-class variance in the feature space for the given class. For this purpose two loss functions, \textit{compactness loss} and \textit{descriptiveness loss} are proposed along with a parallel CNN architecture. A template matching-based framework is introduced to facilitate the testing process.  Extensive experiments on publicly available anomaly detection, novelty detection and mobile active authentication datasets show that the proposed Deep One-Class (DOC) classification method achieves significant improvements over the state-of-the-art.
	
\end{abstract}


\maketitle


	
	\section{Introduction}\label{sec:introduction}

\begin{figure}[t]
	\centering
	\includegraphics[width=1\linewidth]{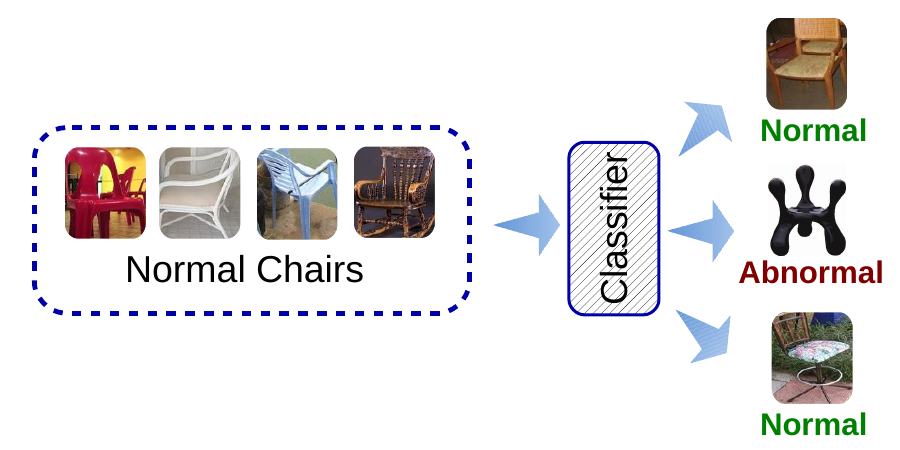}\hskip30pt
\caption{In one class classification, given samples of a single class, a classifier is required to learn so that it can identify out-of-class(alien) objects. Abnormal image detection is an application of one class classification. Here, given a set of normal chair objects, a classifier is learned to detect abnormal chair objects.}
	\label{fig:scales}
\end{figure}

One-class classification is a classical machine learning problem that has received considerable attention in the recent literature\cite{AND},\cite{dsvdd},\cite{ocnn},\cite{Perera_btas_oneclass},\cite{Perera_CVPR19_2},\cite{oza2019one}. The objective of one-class classification is to recognize instances of a concept by only using examples of the same concept \cite{He:2013:ILF:2559492} as shown in Figure~\ref{fig:scales}. In such a training scenario, instances of only a single object class are available during training. In the context of this paper, all other classes except the class given for training are called alien classes.\footnote{Depending on the application, an alien class may be called by an alternative term. For example, an alien class is known as novel class, abnormal class/outlier class, attacker/intruder class in novelty detection, abnormal image detection and active-authentication applications, respectively.} During testing, the classifier may encounter objects from alien classes. The goal of the classifier is to distinguish the objects of the known class from the objects of alien classes. It should be noted that one-class classification is different from binary classification due to the absence of training data from a second class.

One-class classification is encountered in many real-world computer vision applications  including novelty detection \cite{Markou03noveltydetection}, anomaly detection \cite{Chandola:2009:ADS:1541880.1541882}, \cite{roberts}, medical imaging and mobile active authentication  \cite{AA_SPM},\cite{Perera_DAQCD},\cite{oza2019active},\cite{Perera_MAA}. In all of these applications, unavailability of samples from alien classes is either due to the openness of the problem or due to the high cost associated with obtaining the samples of such classes. For example, in a novelty detection application, it is counter intuitive to come up with novel samples to train a classifier. On the other hand, in mobile active authentication, samples of alternative classes (users) are often difficult to obtain due to the privacy concerns \cite{UMDAA02}. 

Despite its importance, contemporary one-class classification schemes trained solely on the given concept have failed to produce promising results in real-world datasets (\cite{AND},\cite{dsvdd} has achieved an Area Under the Curve in the range of 60\%-65\% for CIFAR10 dataset\cite{CIFAR}\cite{Perera_CVPR19_2}). However, we note that computer vision is a field rich with labeled datasets of different domains. In this work, in the spirit of transfer learning, we step aside from the conventional one-class classification protocol and investigate how data from a different domain can be used to solve the one-class classification problem. We name this specific problem \textit{One-class transfer learning} and address it by engineering deep features targeting one-class classification tasks.

\begin{figure*}[t]
	\centering
	\includegraphics[width=1\linewidth]{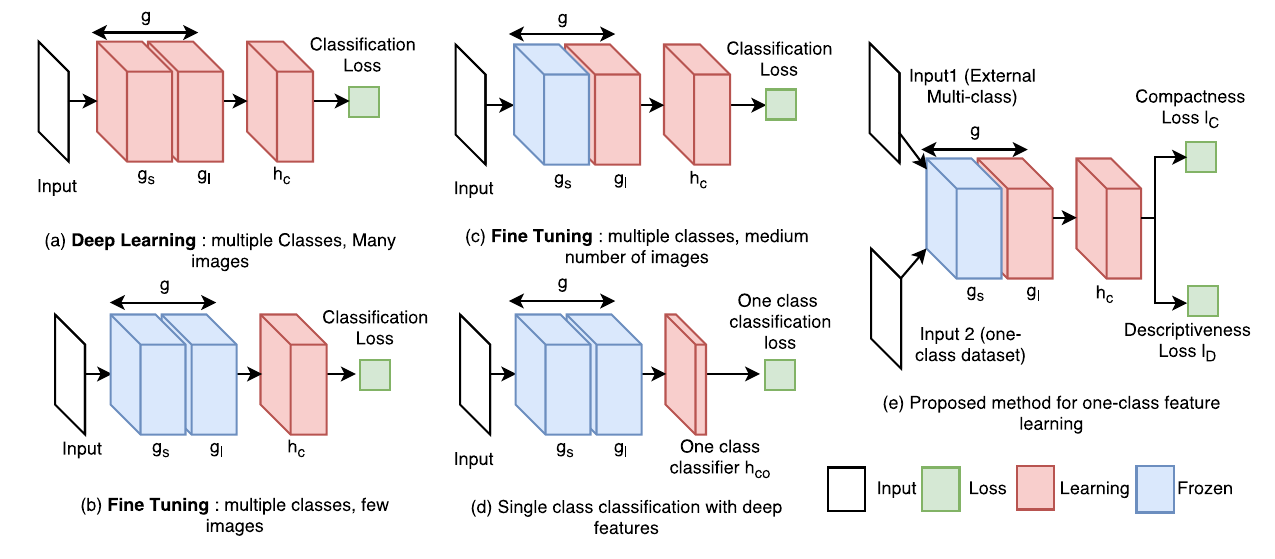}\hskip30pt
	\caption{Different deep learning strategies used for classification. In all cases, learning starts from a pre-trained model. Certain subnetworks (in blue) are frozen while others (in red) are learned during training. Setups (a)-(c) cannot be used directly for the problem of one-class classification.  Proposed method shown in \textbf{(e)} accepts two inputs at a time - one from the given one-class dataset and the other from an external multi-class dataset and produces two losses. }
	\label{fig:over}
\end{figure*}
In order to solve the  \textit{One-class transfer learning} problem, we seek motivation from generic object classification frameworks. Many previous works in object classification have focused on improving either the feature or the classifier (or in some cases both) in an attempt to improve the classification performance. In particular, various deep learning-based feature extraction and classification methods have been proposed  in the literature and have gained a lot of traction in recent years \cite{NIPS2012_ALEX}, \cite{DBLP:journals/corr/SimonyanZ14a}.  In general, deep learning-based classification schemes have two subnetworks, a feature extraction network ($g$) followed by a classification sub network ($h_c$), that are learned jointly during training. For example, in the popular AlexNet architecture \cite{NIPS2012_ALEX}, the collection of convolution layers may be regarded as ($g$) where as fully connected layers may collectively be regarded as ($h_c$). Depending on the output of the classification sub network ($h_c$), one or more losses are evaluated to facilitate training. Deep learning requires the availability of multiple classes for training and extremely large number of training samples (in the order of thousands or millions). However, in learning tasks where either of these conditions are not met, the following alternative strategies are used: 

\noindent \textbf{(a) Multiple classes, many training samples:} This is the case where both requirements are satisfied. Both feature extraction and classification networks, $g$ and $h_c$ are trained end-to-end (Figure~\ref{fig:over}(a)). The network parameters are initialized using random weights. Resultant model is used as the pre-trained model for fine tuning \cite{NIPS2012_ALEX}, \cite{he15deepresidual}, \cite{Perera_CVPR19_1}.

\noindent \textbf{(b) Multiple classes, low to medium number of training samples:} The feature extraction network from a pre-trained model is used. Only a new classification network is trained in the case of low training samples (Figure~\ref{fig:over}(b)). When medium number of training samples are available, feature extraction network ($g$) is divided into two sub-networks - shared feature network ($g_s$) and learned feature network ($g_l$), where $g = g_s \circ g_l$. Here, $g_s$ is taken from a pre-trained model. $g_l$ and the classifier are learned from the data in an end-to-end fashion (Figure~\ref{fig:over}(c)). This strategy is often referred to as fine-tuning \cite{girshick14CVPR}.

\noindent \textbf{(c) Single class or no training data:} A pre-trained model is used to extract features. The pre-trained model used here could be a model trained from scratch (as in (a)) or a model resulting from fine-tuning (as in (b)) \cite{pmlr-v32-donahue14}, \cite{UMDAA02} where training/fine-tuning is performed based on an external dataset. When training data from a class is available, a one-class classifier is trained on the extracted features (Figure~\ref{fig:over}(d)).

 In this work, we focus on the task presented in case (c) where training data from a single class is available. Strategy used in case (c) above uses deep-features extracted from a pre-trained model, where training is carried out on a different dataset, to perform one-class classification. However, there is no guarantee that features extracted in this fashion will be as effective in the new one-class classification task. In this work, we present a feature fine tuning framework which produces deep features that are specialized to the task at hand. Once the features are extracted, they  can be used to perform classification using the strategy discussed in (c).

\begin{figure}[t]
	\centering
	\includegraphics[width=0.75\linewidth]{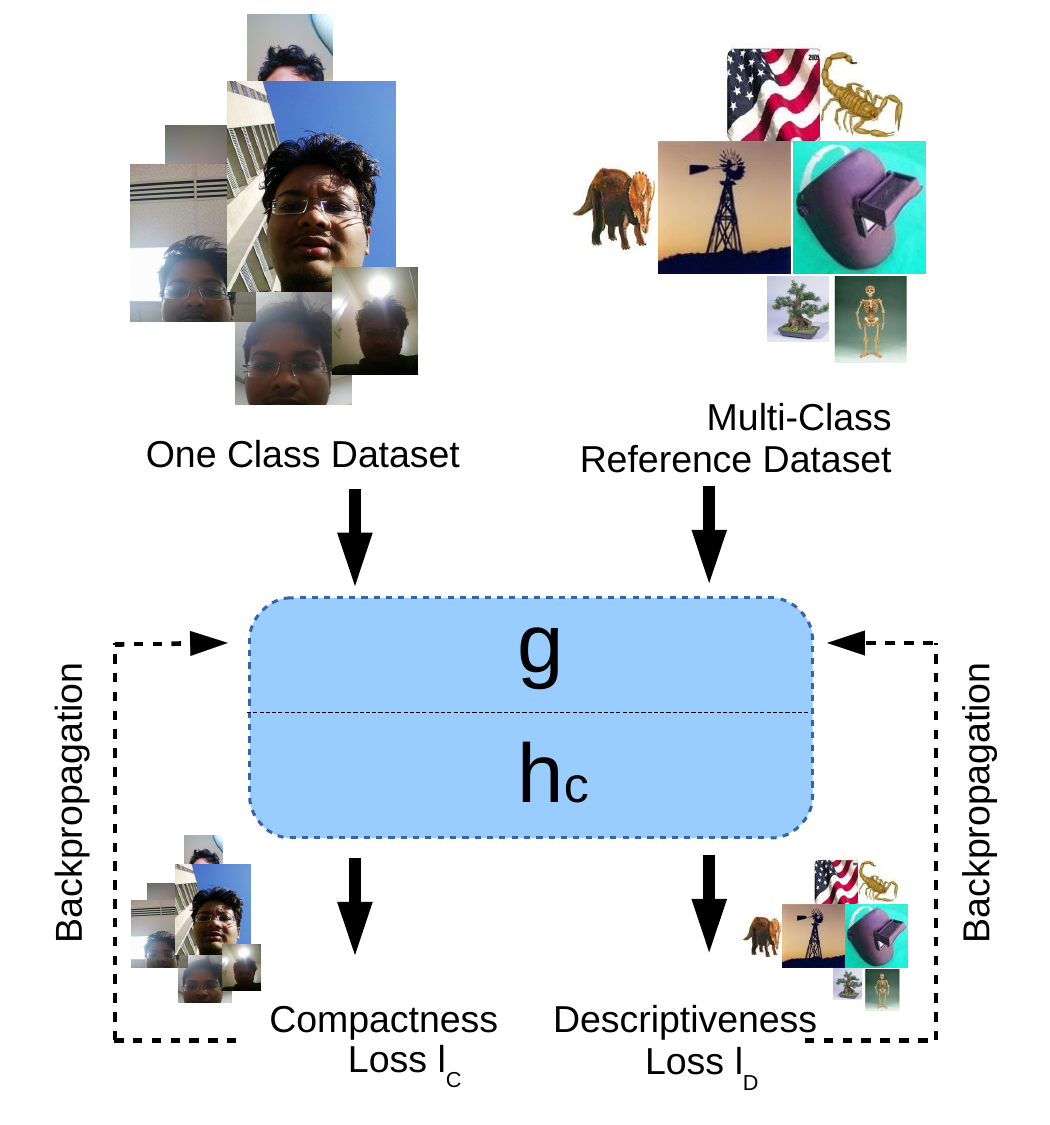}\hskip30pt
	\vskip -0pt\caption{Overview of the proposed method. In addition to the provided one-class dataset, another arbitrary multi-class dataset is used to train the CNN. The objective is to produce a CNN which generates descriptive features which are also compactly localized for the one-class training set. Two losses are evaluated independently with respect to the two datasets and back-propagation is performed to facilitate training.}
	\label{fig:fig:overview}
\end{figure}

In our formulation (shown in Figure~\ref{fig:over} (e)), starting from a pre-trained deep model, we freeze initial features ($g_s$) and learn ($g_l$) and ($h_c$). Based on the output of the classification sub-network ($h_c$), two losses \textit{compactness loss} and \textit{descriptiveness loss} are evaluated.  These two losses, introduced in the subsequent sections, are used to assess the quality of the learned deep feature. We use the provided one-class dataset to calculate the \textit{compactness loss}. An external multi-class reference dataset is used to evaluate the \textit{descriptiveness loss}. As shown in Figure~\ref{fig:fig:overview}, weights of  $g_l$ and $h_c$  are learned in the proposed method through back-propagation from the composite loss. Once training is converged, system shown in setup in Figure~\ref{fig:over}(d) is used to perform classification where the resulting model is used as the pre-trained model. 

In summary, this paper makes the following three contributions.
\begin{enumerate}
\item We propose a deep-learning based feature engineering scheme called Deep One-class Classification (DOC), targeting one-class problems. To the best of our knowledge this is one of the first works to address this problem.

\item We introduce a joint optimization framework based on two loss functions - \textit{compactness loss} and \textit{descriptiveness loss}. We propose \textit{compactness loss} to assess the compactness of the class under consideration in the learned feature space. We propose using an external multi-class dataset to assess the descriptiveness of the learned features using \textit{descriptiveness loss}.

\item On three publicly available datasets, we achieve state-of-the-art one-class classification performance across three different tasks. 
\end{enumerate}

Rest of the paper is organized as follows.  In Section~\ref{sec:related}, we review a few related works.  Details of the proposed deep one-class classification method are given in Section~\ref{sec:method}.  Experimental results are presented in Section~\ref{sec:results}.  Section~\ref{sec:con} concludes the paper with a brief discussion and summary.

\section{Related Work}\label{sec:related}
Generic one-class classification problem has been addressed using several different approaches in the literature.  Most of these approaches predominately focus on proposing a better classification scheme operating on a chosen feature and are not specifically designed for the task of one-class image classification. One of the initial approaches to one-class learning was to estimate a parametric generative model based on the training data. Work in \cite{DBLP:conf/visapp/2009-s}, \cite{Bishop:2006:PRM:1162264}, propose to use Gaussian distributions and Gaussian Mixture Models (GMMs) to represent the underlying distribution of the data. In \cite{Kemmler:2013:OCG:2514172.2514288} comparatively better performances have been obtained by estimating the conditional density of the one-class distribution using Gaussian priors.

The concept of Support Vector Machines (SVMs) was extended for the problem of one-class classification in \cite{Scholkopf:2001:ESH:1119748.1119749}.  Conceptually this algorithm treats the origin as the out-of-class region and tries to construct a hyperplane separating the origin with the class data.  Using a similar motivation,  \cite{Tax:2004:SVD:960091.960109}  proposed Support Vector Data Description (SVDD) algorithm which isolates the training data by constructing a spherical separation plane. In \cite{ELM}, a single layer neural network based on Extreme Learning Machine is proposed for one-class classification. This formulation results in an efficient optimization as layer parameters are updated using closed form equations.  Practical one-class learning applications on different domains are predominantly developed based on these conceptual bases. 

In \cite{Rodner11one-classclassification}, visual anomalies in wire ropes are detected based on Gaussian process modeling. Anomaly detection is performed by maximizing KL divergence  in \cite{rodner:2016}, where the underlying distribution is assumed to be a known Gaussian. A detailed description of various anomaly detection methods can be found in \cite{Chandola:2009:ADS:1541880.1541882}.

Novelty detection based on one-class learning has also received a significant attention in recent years. Some of the earlier works in novelty detection focused on estimating a parametric model for data and to model the tail of the distribution to improve classification accuracy \cite{clifton2011novelty},\cite{roberts}. In \cite{Bodesheim_2013_CVPR}, null space-based novelty detection framework for scenarios when a single and multiple classes are present is proposed. However, it is mentioned in \cite{Bodesheim_2013_CVPR} that their method does not yield superior results compared with the classical one-class classification methods when only a single class is present. An alternative null space-based approach based on kernel Fisher discriminant was proposed in \cite{DBLP:conf/ijcnn/2016} specifically targeting one-class novelty detection. A detailed survey of different novelty detection schemes can be found in \cite{Markou03noveltydetection}, \cite{Markou:2003:NDR:959414.959416}.

Mobile-based Active Authentication (AA) is another application of one-class learning which has gained interest of the research community in recent years \cite{AA_SPM}. In mobile AA, the objective is to continuously monitor the identity of the user based on his/her enrolled data. As a result, only the enrolled data (i.e. one-class data) are available during training. Some of the recent works in AA has taken advantage of CNNs for classification. Work in \cite{Pouya_BTAS2016}, uses a CNN to extract attributes from face images extracted from the mobile camera to determine the identity of the user. Various deep feature-based AA methods have also been proposed as benchmarks in \cite{UMDAA02} for performance comparison.

Since one-class learning is constrained with training data from only a single class, it is impossible to adopt a CNN architectures used for classification \cite{NIPS2012_ALEX}, \cite{DBLP:journals/corr/SimonyanZ14a} and verification  \cite{Chopra:2005:SIAMESE} directly for this problem.  In the absence of a discriminative feature generation method, in most unsupervised tasks, the activation of a deep layer is used as the feature for classification. This approach is seen to generate satisfactory results in most applications \cite{pmlr-v32-donahue14}. This can be used as a starting point for one-class classification as well. As an alternative, autoencoders \cite{Vincent:2008:AE}, \cite{MNISTAUTO} and variants of autoencoders \cite{Vincent:2010:SDA:1756006.1953039}, \cite{Variational_autoencoders} can also to be used as feature extractors for one-class learning problems. However, in this approach, knowledge about the outside world is not taken into account during the representation learning. Furthermore, none of these approaches were specifically designed for the purpose of one-class learning. To the best of our knowledge, one-class feature learning has not been addressed using a deep-CNN architecture to date.

\section{Deep One-class Classification (DOC)}\label{sec:method}
\subsection{Overview}

In this section, we formulate the objective of one-class feature learning as an optimization problem. In the classical multiple-class classification, features are learned with the objective of maximizing inter-class distances between classes and minimizing intra-class variances within classes \cite{BelhumeurHK97}. However, in the absence of multiple classes such a discriminative approach is not possible.

In this light, we outline two important characteristics of features intended for one-class classification. 

\noindent \textbf{Compactness $\mathcal{C}$.} 
A desired quality of a feature is to have a similar feature representation for different images of the same class. Hence, a collection of features extracted from a set of images of a given class will be compactly placed in the feature space. This quality is desired even in features used for multi-class classification. In such cases, compactness is quantified using  the intra-class distance  \cite{BelhumeurHK97}; a compact representation would have a lower intra-class distance. 

\noindent \textbf{Descriptiveness $\mathcal{D}$.} The given feature should produce distinct representations for images of different classes. Ideally, each class will have a distinct feature representation from each other. Descriptiveness in the feature is also a desired quality in multi-class classification. There, a descriptive feature would have large inter-class distance  \cite{BelhumeurHK97}.

It should be noted that for a useful (discriminative) feature, both of these characteristics should be satisfied collectively. Unless mutually satisfied, neither of the above criteria would result in a useful feature. With this requirement in hand, we aim to find a feature representation $g$ that maximizes both compactness and descriptiveness. Formally, this can be stated as an optimization problem as follows,
\begin{equation} \label{eqn:opt}
\hat{g}=\max_{g}~  \mathcal{D} (g(t)) + \lambda \mathcal{C}(g(t)),
\end{equation}
where $t$ is the training data corresponding to the given class and $\lambda$ is a positive constant. Given this formulation, we identify three potential strategies that may be employed when deep learning is used for one-class problems. However, none of these strategies collectively satisfy both descriptiveness and compactness.

\noindent \textbf{(a) Extracting deep features.} Deep features are first extracted from a pre-trained deep model for given training images. Classification is done using a one-class classification method such as one-class SVM, SVDD or k-nearest neighbor using extracted features. This approach does not directly address the two characteristics of one-class features. However, 
if the pre-trained model used to extract deep features was trained on a dataset with large number of classes, then resulting deep features are likely to be descriptive. Nevertheless, there is no guarantee that the used deep feature will possess the compactness property.

\noindent \textbf{(b) Fine-tune a two class classifier using an external dataset.} Pre-trained deep networks are trained based on some legacy dataset. For example, models used for the ImageNet challenge are trained based on the ImageNet dataset \cite{imagenet_cvpr09}. It is possible to fine tune the model by representing the alien classes using the legacy dataset. This strategy will only work when there is a high correlation between  alien classes and the legacy dataset. Otherwise, the learned feature will not have the capacity to describe the difference between a given class and the alien class thereby violating the descriptiveness property. 

\noindent \textbf{(c) Fine-tune using a single class data.} Fine-tuning may be attempted by using  data only from the given single class. For this purpose, minimization of the traditional cross-entropy loss or any other appropriate distance could be used. However, in such a scenario, the network may end up learning a trivial solution due to the absence of a penalty for miss-classification. In this case, the learned representation will be compact but will not be descriptive.

\begin{figure*}[t]
	\centering
	\includegraphics[width=\linewidth]{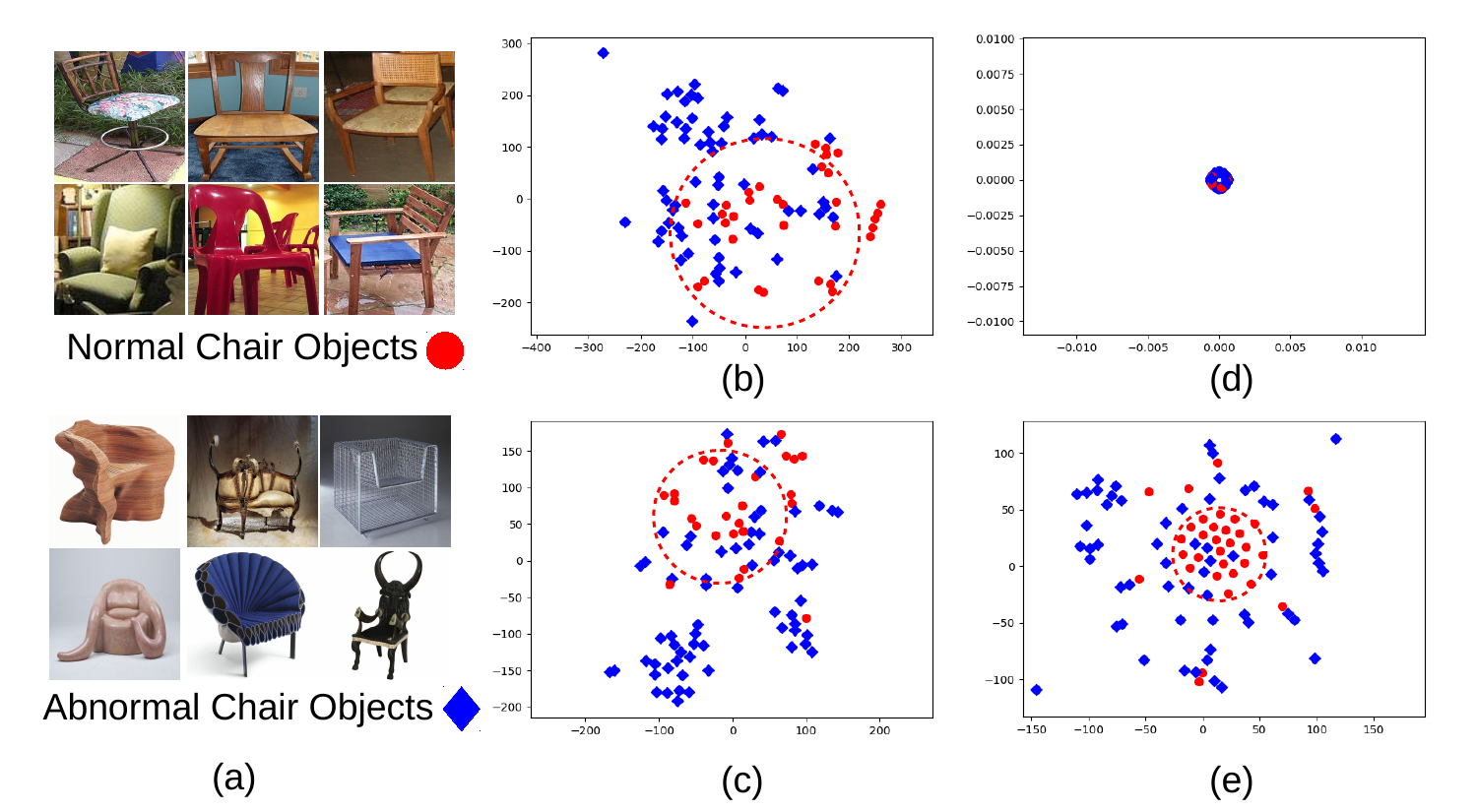}\hskip30pt
	\caption{Possible strategies for one-class classification in abnormal image detection. (a) Normal and abnormal image samples. (b) Feature space  obtained using the AlexNet features. (c) Feature space obtained training a two class CNN with the alien objects represented by the ImageNet data samples. Normal and abnormal samples are not sufficiently separated in (b) and (c). (d) Feature space (2D Projection) obtained by fine-tuning using  just normal objects. Both normal and abnormal samples are projected to the same point. (e) Feature space (2D Projection) obtained using the proposed method. Normal and abnormal samples are relatively more separated. In all cases, feature space is visualized using tSNE. SVDD decision boundary for the normal class is drawn in dotted lines in the tSNE space.}
	\label{fig:features}
\end{figure*}

Let us investigate the appropriateness of these three strategies by conducting a case study on the abnormal image detection problem where the considered class is the \textit{normal chair} class. In abnormal image detection, initially a set of  \textit{normal chair} images are provided for training as shown in Figure~\ref{fig:features}(a). The goal is to learn a representation such that, it is possible to distinguish a \textit{normal chair} from an \textit{abnormal chair}. 

The trivial approach to this problem is to extract deep features from an existing CNN architecture (solution (a)). Let us assume that the AlexNet architecture \cite{NIPS2012_ALEX} is used for this purpose and \textit{fc7} features are extracted from each sample. Since deep features are sufficiently descriptive, it is reasonable to expect samples of the same class to be clustered together in the extracted feature space. Illustrated in Figure~\ref{fig:features}(b) is a 2D visualization of the extracted 4096 dimensional features using t-SNE \cite{ictdbid:2777}. As can be seen from Figure\ref{fig:features}(b), the AlexNet features are not able to enforce sufficient separation between normal and abnormal chair classes.

Another possibility is to train a two class classifier using the AlexNet architecture by providing normal chair object images and the images from the ImageNet dataset as the two classes (solution (b)). However, features learned in this fashion produce similar representations for both normal and abnormal images, as shown in Figure\ref{fig:features}(c). Even though there exist subtle differences between normal and abnormal chair images, they have more similarities  compared to the other ImageNet objects/images. This is the main reason why both normal and abnormal images end up learning similar feature representations.

A naive, and ineffective, approach would be to fine-tune the pre-trained AlexNet network using only the \textit{normal chair} class (solution (c)). Doing so, in theory, should result in a representation where all \textit{normal chairs} are compactly localized in the feature space. However, since all class labels would be identical in such a scenario, the fine-tuning process would end up learning a futile representation as shown in Figure\ref{fig:features}(d). The reason why this approach ends up yielding a trivial solution is due to the absence of a regularizing term in the loss function that takes into account the discriminative ability of the network. For example, since all class labels are identical, a zero loss can be obtained by making all weights equal to zero. It is true that this is a valid solution in the closed world where only \textit{normal chair} objects exist.  But such a network has zero discriminative ability when \textit{abnormal chair} objects appear. 

None of the three strategies discussed above are able to produce features that are both compact and descriptive. We note that out of the three strategies, the first produces the most reasonable representation for one-class classification. However, this representation was learned without making an attempt to increase compactness of the learned feature. Therefore, we argue that if compactness is taken into account along with descriptiveness, it is possible to learn a more effective representation. 

\subsection{Proposed Loss Functions}
In this work, we propose to quantify compactness and descriptiveness in terms of measurable loss functions. Variance of a distribution has been widely used in the literature as a measure of the distribution spread \cite{Gubner:2006:PRP:1207291}. Since spread of the distribution is inversely proportional to the compactness of the distribution, it is a natural choice to use variance of the distribution to quantify compactness. In our work, we approximate variance of the feature distribution by the variance of each feature batch. We term this quantity as the \textit{compactness loss $(l_C)$}.

On the other hand, descriptiveness of the learned feature cannot be assessed using a single class training data. However, if there exists a reference dataset with multiple classes, even with random object classes unrelated to the problem at hand, it can be used to assess the descriptiveness of the engineered feature. In other words, if the learned feature is able to perform classification with high accuracy on a different task, the descriptiveness of the learned feature is high. Based on this rationale, we use the learned feature to perform classification on an external multi-class dataset, and consider classification loss there as an indicator of the descriptiveness of the learned feature. We call the cross-entropy loss calculated in this fashion as the \textit{descriptiveness loss $(l_D)$}. Here, we note that \textit{descriptiveness loss} is low for a descriptive representation.


With this formulation, the original optimization objective in equation~\eqref{eqn:opt} can be re-formulated as,
\begin{equation} \label{eqn:opt2}
\hat{g}=\min_{g}~  l_D(r) + \lambda l_C(t),
\end{equation}
where $l_C$ and $l_D$ are \textit{compactness loss} and \textit{descriptiveness loss}, respectively and $r$ is the training data corresponding to the reference dataset. The tSNE visualization of the features learned in this manner for normal and abnormal images are shown in Figure~\ref{fig:features}(e). Qualitatively, features learned by the proposed method facilitate better distinction between normal and abnormal images as compared with the cases is shown in Figure~\ref{fig:over}(b)-(d).

\subsection{Terminology}
\begin{figure*}[t]
	\centering
	\includegraphics[width=\linewidth]{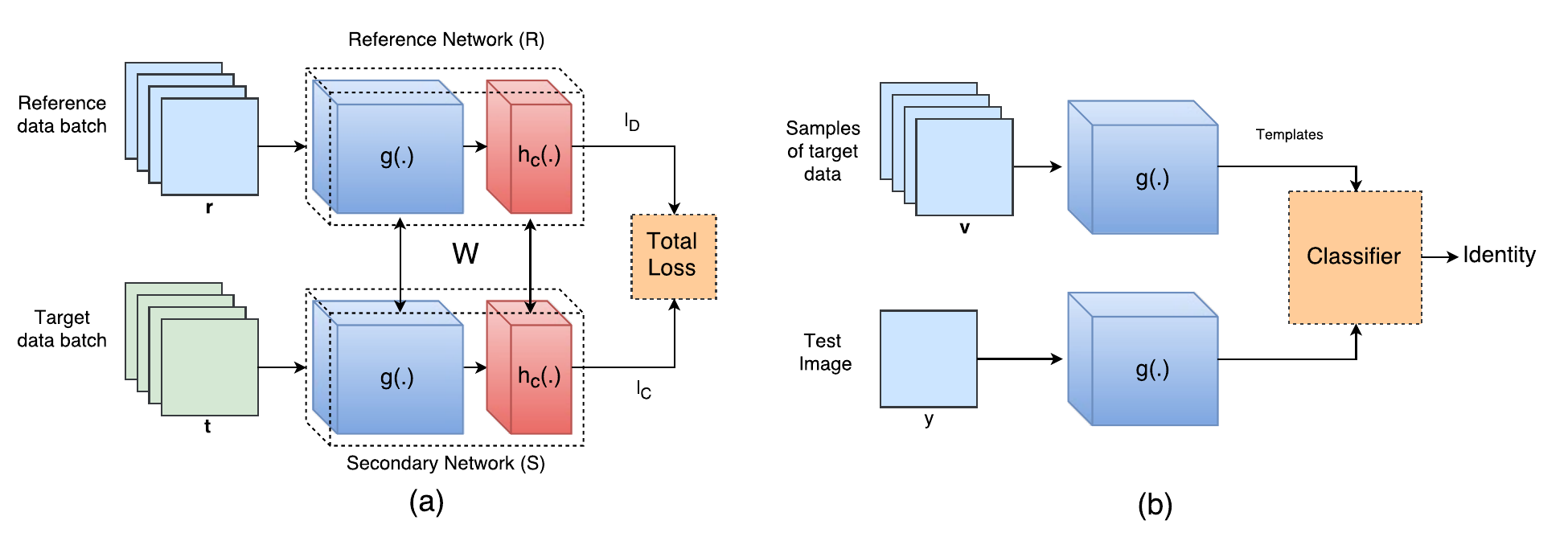}\hskip30pt
	\vskip -0pt\caption{(a) Training, and (b) testing frameworks of the proposed DOC method. During training, data from target and reference datasets are fed into the network simultaneously and training is done based on two losses, \textit{descriptive loss} and \textit{compactness loss}. During testing, only the feature extraction sub-network $g$ is considered. A set of templates is first constructed by extracting features of a subset of the training dataset. When a test image is present, features of the test image is compared against templates using a classification rule to classify the test image.}
	\label{fig:arch}
\end{figure*}

Based on the intuition given in the previous section, the architecture shown in Figure~\ref{fig:arch} (a) is proposed for one-class classification training and the setup shown in Figure~\ref{fig:arch} (b) for testing. They consist of following elements:

\noindent {\bf{Reference Network ($R$):}}
This is a pre-trained network architecture considered for the application. Typically it contains a repetition of convolutional, normalization, and pooling layers (possibly with skip connections) and is terminated by an optional set of fully connected layers. For example, this could be the AlexNet network \cite{NIPS2012_ALEX} pre-trained using the ImageNet \cite{imagenet_cvpr09} dataset. Reference network can be seen as the composition of a feature extraction sub-network $g$ and a classification sub-network $h_{c}$. For example, in AlexNet, \textit{conv1-fc7} layers can be associated with $g$ and fc8 layer with $h_{c}$. \textit{Descriptive loss} $(l_D)$ is calculated based on the output of $h_{c}$.

\noindent {\bf{Reference Dataset} ($\boldmath{r}$):}
This is the dataset (or a subset of it) used to train the network $R$. Based on the example given, reference dataset is the ImageNet dataset \cite{imagenet_cvpr09} (or just a subset of the ImageNet dataset).

\noindent {\bf{Secondary Network ($S$):}} This is a second CNN where the network architecture is structurally identical to the reference network. Note that $g$ and $h_c$ are shared in both of these networks. \textit{Compactness loss} $(l_C)$ is evaluated based on the output of $h_c$. For the considered example, $S$ would have the same structure as  $R$ (AlexNet) up to \textit{fc8}.

\noindent {\bf{Target Dataset  }($\boldmath{t}$):} This dataset contains samples of the class for which one-class learning is used for. For example, for an abnormal image detection application, this dataset will contain normal images (i.e. data samples of the single class considered).

\noindent {\bf{Model ($\mathbf{W}$):}} This corresponds to the collection of weights and biases in the network, $g$ and $h_c$. Initially, it is initialized by some pre-trained parameters $W_0$. \footnote{For the case of AlexNet, pre-trained model can be found at www.berkleyvison.com.} 

\noindent {\bf{Compactness loss ($l_C$) :}} All the data used during the training phase will belong to the same class. Therefore they all share the same class label. \textit{Compactness loss} evaluates the average similarity between the constituent samples of a given batch. For a large enough batch, this quantity can be expected to represent average intra-class variance of a given class. It is desirable to select a smooth differentiable function as $l_C$ to facilitate back propagation. In our work, we define compactness loss based on the Euclidean distance.

\noindent {\bf{Descriptiveness loss ($l_D$) :}} Descriptiveness loss evaluates the capacity of the learned feature to describe different concepts. We propose to quantify discriminative loss by  the evaluation of cross-entropy with respect to the reference dataset $(R)$.

 For this discussion, we considered the AlexNet CNN architecture as the reference network. However, the discussed principles and procedures would also apply to any other CNN of choice. In what follows, we present the implementation details of the proposed method. 

\subsection{Architecture} \label{sec:arch}
The proposed training architecture is shown in Figure~\ref{fig:arch} (a) \footnote{Source code of the proposed method is made available online at https://github.com/PramuPerera/DeepOneClass}. The architecture consists of two CNNs, the reference network $(R)$ and the secondary network $(S)$ as introduced in the previous sub-section. Here, weights of the reference network and secondary network are tied across each corresponding counterparts. For example, weights between conv$i$ (where, $i = 1,2..,5$) layer of the two networks are tied together forcing them to be identical. All components, except \textit{Compactness loss}, are standard CNN components. We denote the common feature extraction sub-architecture by $g(.)$ and the common classification by sub-architecture by $h_c(.)$. Please refer to Appendix for more details on the architectures of the proposed method based on the AlexNet and VGG16 networks.

\subsection{Compactness loss}
\textit{Compactness loss} computes the mean squared intra-batch distance within a given batch. In principle, it is possible to select any distance measure for this purpose. In our work, we design compactness loss based on the Euclidean distance measure. Define $\mathbf{X} = \{\mathbf{x_1},\mathbf{x_2}, \dots, \mathbf{x_n}\}\in{R}^{n \times k}$ to be the input to the loss function, where the batch size of the input is $n$.

\noindent {\bf{Forward Pass:}}
For each $i^{th}$ sample $\mathbf{x_i} \in \mathbb{R}^{k}$, where $1 \leq i\leq n$, the distance between the given sample and the rest of the samples  $\mathbf{z_i}$ can be defined as,

\begin{equation} \mathbf{z_i}  =  \mathbf{x_i} - \mathbf{m_i} ,\end{equation}
where, $ \mathbf{m_i} = \frac{1}{n-1}\sum_{j \neq i}\mathbf{x_j}$ is the mean of rest of the samples. Then, compactness loss $l_C$ is defined as the average Euclidean distance as in,

\begin{equation} \label{eq:lb} l_C = \frac{1}{nk}  \sum_{i=1}^{n} \mathbf{z_i}^T\mathbf{z_i}.\end{equation}

\noindent {\bf{Backpropagation:}}. In order to perform back-propagation using this loss, it is necessary to assess the contribution each element of the input has on the final loss. Let $\mathbf{x_i} = \{ x_{i1}, x_{i2}, \dots , x_{ik}\}$. Similarly, let $\mathbf{m_i} = \{ m_{i1}, m_{i2}, \dots , m_{ik}\}$. Then, the gradient of $l_b$ with respect to the input $x_{ij}$ is given as,

\begin{equation} \frac{\partial l_C}{\partial {x_{ij}}} = \frac{2}{(n-1)nk} \bigg[ n \times (x_{ij} - m_{ij}) - \sum_{k=1}^{n} (x_{ik} - m_{ik}) \bigg].\end{equation}  

 Detailed derivation of the back-propagation formula can be found in the Appendix. The loss $l_C$ calculated in this form is equal to the sample feature variance of the batch multiplied by a constant (see Appendix). Therefore, it is an inverse measure of the compactness of the feature distribution.

\subsection{Training}
During the training phase, initially, both the reference network $(R)$ and the secondary network $(S)$ are initialized with the pre-trained model weights $\mathbf{W_0}$. Recall that except for the types of loss functions associated with the output, both these networks are identical in structure. Therefore, it is possible to initialize both networks with identical weights. During training, all layers of the network except for the last four layers are frozen as commonly done in network fine-tuning. In addition, the learning rate of the training process is set at a relatively lower value ( $5 \times 10^{-5}$ is used in experiments). During training, two image batches, each from the reference dataset and the target dataset are simultaneously fed into the input layers of the reference network and secondary network, respectively. At the end of the forward pass, the reference network generates a \textit{descriptiveness loss} ($l_{D}$), which is same as the cross-entropy loss by definition, and the secondary network generates \textit{compactness loss} ($l_C$). The composite loss ($l$) of the network is defined as,
\begin{equation} l(r,t) = l_{D}({r|W}) + \lambda l_C(t|W),\end{equation}
where $\lambda$ is a constant. It should be noted that, minimizing this loss function leads to the minimization of the optimization objective in \eqref{eqn:opt2}.

\begin{figure}[htp!]
	\centering
	\includegraphics[width=1\linewidth]{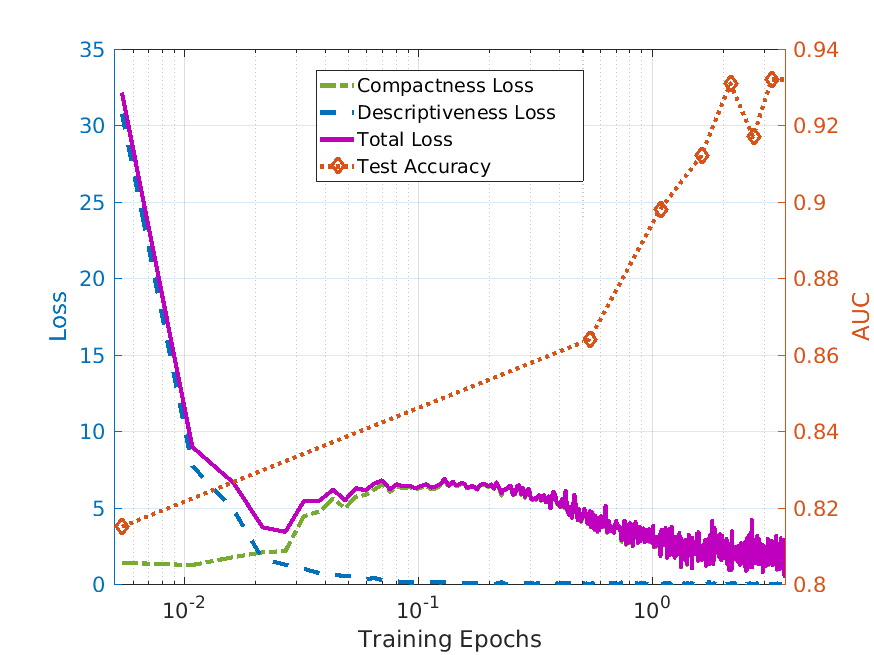}\hskip30pt
	\vskip -0pt\caption{Variation of loss functions during training in the proposed method obtained for the chair class in the abnormal image detection task. Training epochs are shown in the log scale. Compactness loss reaches stability around 0.1 epochs. Total loss (composite loss) reaches stability around 2 epochs. Highest test accuracy is obtained once composite loss converges.    }
	\label{fig:output_trainingLoss}
\end{figure}

In our experiments, $\lambda$ is set equal to 0.1 Based on the composite loss, the network is back-propagated and the network parameters are learned using gradient descent or a suitable variant. Training is carried out until composite loss $l(r,t)$ converges. A sample variation of training loss is shown in Figure~\ref{fig:output_trainingLoss}. In general, it was observed that composite loss converged in around two epochs (here, epochs are defined based on the size of the target dataset).

Intuitively, the two terms of the loss function $l_{D}$ and $l_C$ measure two aspects of features that are useful for one-class learning.  Cross entropy loss values obtained in calculating \textit{descriptiveness loss} $l_{D}$ measures the ability of the learned feature to describe different concepts with respect to the reference dataset. Having reasonably good performance in the reference dataset implies that the learned features are discriminative in that domain. Therefore, they are likely to be descriptive in general. On the other hand, \textit{compactness loss} ($l_C$) measures how compact the class under consideration is in the learned feature space. The weight $\lambda$ governs the mutual importance placed on each requirement. 

If $\lambda$ is made large, it implies that the descriptiveness of the feature is not as important as the compactness. However, this is not a recommended policy for one-class learning as doing so would result in trivial features where the overlap between the given class and an alien class is significantly high. As an extreme case, if $\lambda = 0$ (this is equivalent to removing the reference network and carrying out training solely on the secondary network (Figure~\ref{fig:over} (d)), the network will learn a trivial solution. In our experiments, we found that in this case the weights of the learned filters become zero thereby making output of any input equal to zero. 

Therefore, for practical one-class learning problems, both reference and secondary networks should be present and more prominence should be given to the loss of the reference network.

\subsection{Testing} 
The proposed testing procedure involves two phases - template generation and matching. For both phases, secondary network with weights learned during training is used as shown in Figure~\ref{fig:arch} (b). During both phases, the excitation map of the feature extraction sub-network is used as the feature. For example, layer 7, \textit{fc7} can be considered from a AlexNet-based network. First, during the template generation phase a small set of samples $\boldmath{v} = \{v_1,v_2, \dots ,v_n  \}$ are drawn from the target (i.e. training) dataset where $\boldmath{v} \in \boldmath{t}$. Then, based on the drawn samples a set of features $g(v_1),g(v_2), \dots ,g(v_n) $ are extracted. These extracted features are stored as  templates and will be used in the matching phase. 

Based on stored template, a suitable one-class classifier, such as one-class SVM \cite{Scholkopf:2001:ESH:1119748.1119749}, SVDD \cite{Tax:2004:SVD:960091.960109} or k-nearest neighbor, can be trained on the templates. In this work, we choose the simple k-nearest neighbor classifier described below. When a test image $y$ is present, the corresponding deep feature $g(y)$ is extracted using the described method. Here, given a set of templates, a matched score $S_y$ is assigned to $y$ as 
\begin{equation}S_y = f(g(y)|g(t_1),g(t_2),\dots,g(t_n)), \end{equation}
where $f(.)$ is a matching function. This matching function can be a simple rule such as the cosine distance or it could be a more complicated function such as Mahalanobis distance. In our experiments, we used Euclidean distance as the matching function.  After evaluating the matched score, $y$ can be classified as follows, 
\begin{equation}
class(y) = 
\begin{cases}
1, &  \text{if } S_y \leq \delta\\
0,      &  \text{if }  S_y > \delta,
\end{cases}
\end{equation}
where 1 is the class identity of the class under consideration and 0 is the identity of other classes and $\delta$ is a threshold. 	

\subsection{Memory Efficient Implementation}
Due to shared weights between the reference network and the secondary network, the amount of memory required to store the network is nearly  twice as the number of parameters. It is not possible to take advantage of this fact with deep frameworks with static network architectures (such as caffe \cite{caffe}). However, when frameworks that support dynamic network structures are used (e.g. PyTorch), implementation can be altered to reduce memory consumption of the network.

In the alternative implementation, only a single core network with functions $g$  and $h_c$ is used. Two loss functions $l_C$ and $l_D$ branch out from the core network. However in this setup, \textit{descriptiveness loss ($l_D$)} is scaled by a factor of $1-\lambda$. In this formulation, first $\lambda$ is made equal to 0 and a data batch from the reference dataset is fed into the network. Corresponding loss is calculated and resulting gradients are calculated using back-propagation Then, $\lambda$ is made equal to 1 and a data batch from the target dataset is fed into the network. Gradients are recorded same as before after back-propagation. Finally, the average gradient is calculated using two recorded gradient values, and network parameters are updated accordingly. In principle, despite of having a lower memory requirement, learning behavior in the alternative implementation would be identical to the original formulation.

\section{Experimental Results}\label{sec:results}

In order to asses the effectiveness of the proposed method, we consider three one-class classification tasks: abnormal image detection, single class image novelty detection and active authentication. We evaluate the performance of the proposed method in all three cases against state of the art methods using publicly available datasets. Further, we provide two additional CNN-based baseline comparisons.
 
\subsection{Experimental Setup}
Unless otherwise specified, we used 50\% of the data for training and the remaining data samples for testing. In all cases, 40 samples were taken at random from the training set to generate templates. In datasets with multiple classes, testing was done by treating one class at a time as the positive class. Objects of all the other classes were considered to be alien. During testing, alien object set was randomly sampled to arrive at equal number of samples as the positive class. As for the reference dataset, we used the validation set of the ImageNet dataset for all tests. When there was an object class overlap between the target dataset and the reference dataset, the corresponding overlapping classes were removed from the reference dataset. For example, when novelty detection was performed based on the Caltech 256, classes appearing in both Caltech 256 and ImageNet were removed from the ImageNet dataset prior to training. 

The Area Under the Curve (AUC) of the Receiver Operating Characteristic (ROC) Curve are used to measure the performance of different methods.   The reported performance figures in this paper are the average AUC figures obtained by considering multiple classes available in the dataset. In all of our experiments, Euclidean distance was used to evaluate the similarity between a test image and the stored templates. In all experiments, the performance of the proposed method was evaluated based on both the AlexNet \cite{NIPS2012_ALEX} and the VGG16 \cite{DBLP:journals/corr/SimonyanZ14a} architectures. In all experimental tasks, the following experiments were conducted.  

\noindent \textbf{AlexNet Features and VGG16 Features (Baseline). } One-class classification is performed using k-nearest neighbor, One-class SVM\cite{Scholkopf:2001:ESH:1119748.1119749}, Isolation Forest\cite{Bishop:2006:PRM:1162264} and Gaussian Mixture Model\cite{Bishop:2006:PRM:1162264} classifiers on $fc7$ AlexNet features and the $fc7$ VGG16 features, respectively.

\noindent \textbf{AlexNet Binary and VGG16 Binary (Baseline).} A binary CNN is trained by having ImageNet samples and one-class image samples as the two classes using AlexNet and VGG16 architectures, respectively. Testing is performed using k-nearest neighbor, One-class SVM\cite{Scholkopf:2001:ESH:1119748.1119749}, Isolation Forest\cite{Bishop:2006:PRM:1162264} and Gaussian Mixture Model\cite{Bishop:2006:PRM:1162264} classifiers.

\noindent \textbf{One-class Neural Network (OCNN).} Method proposed in \cite{ocnn} applied on the extracted features from the AlexNet and VGG16 networks.

\noindent \textbf{Autoencoder \cite{MNISTAUTO}.} Network architecture proposed in \cite{MNISTAUTO} is used to learn a representation of the data. Reconstruction loss is used to perform verification.

\noindent \textbf{Ours (AlexNet) and ours (VGG16).} Proposed method applied with AlexNet and VGG16 network backbone architectures. The $fc7$ features are used during testing.

In addition to these baselines, in each experiment we report the performance of other task specific methods.

\subsection{Results}

\begin{figure}[t]
	\centering
	\includegraphics[width=1\linewidth]{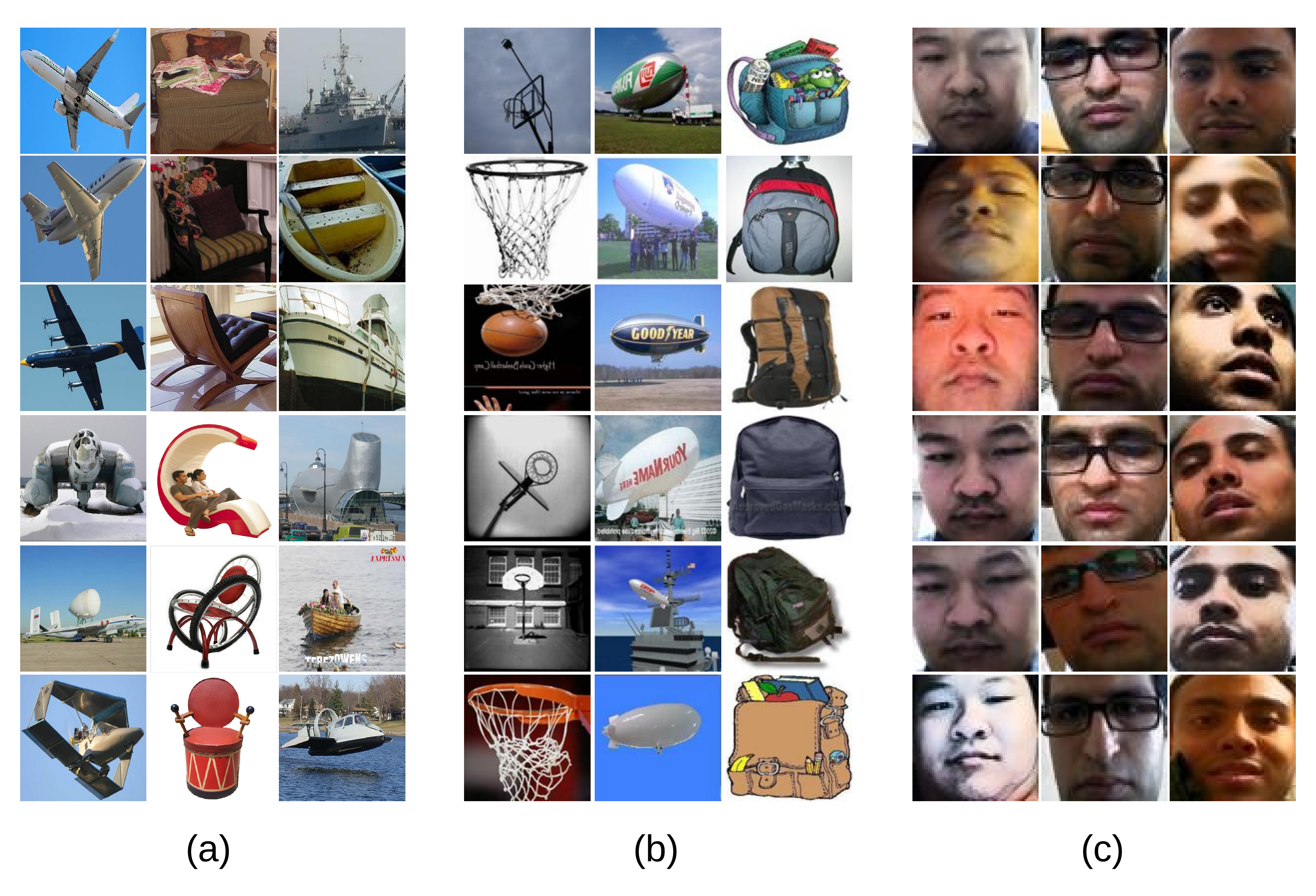}\hskip30pt
	\caption{Sample images from datasets used for evaluation. (a) Abnormal image detection : Top three rows are normal images taken from PASCAL dataset. Bottom three rows are abnormal images taken from Abnormal 1001 dataset. (b) Novelty detection: images from Caltech 256 dataset. (c) Active Authentication: images taken from UMDAA02 dataset. In (b) and (c) columns are formed by images from a single class.}
	\label{fig:samples}
\end{figure}

\noindent {\bf{Abnormal Image Detection:}}
The goal in abnormal image detection is to detect abnormal images when the classifier is trained using a set of normal images of the corresponding class. Since the nature of abnormality is unknown a priori, training is carried out using a single class (images belonging to the normal class). The 1001 Abnormal Objects Dataset \cite{Saleh:2013:OAD:2514950.2516141} contains 1001 abnormal images belonging to six classes which are originally found in the PASCAL \cite{Everingham10} dataset.  Six classes considered in the dataset are Airplane, Boat, Car, Chair, Motorbike and Sofa. Each class has at least one hundred abnormal images in the dataset. A sample set of abnormal images and the corresponding normal images in the PASCAL dataset are show in Figure~\ref{fig:samples}(a). Abnormality of images has been judged based on human responses received on the Amazon Mechanical Turk. We compare the performance of abnormal  detection of the proposed framework with conventional CNN schemes and with the comparisons presented in \cite{Saleh:2013:OAD:2514950.2516141}. It should be noted that our testing procedure is consistent with the protocol used in \cite{Saleh:2013:OAD:2514950.2516141}.

Results corresponding to this experiment are shown in Table~\ref{table:ab1001}. Adjusted graphical model presented in \cite{Saleh:2013:OAD:2514950.2516141} has outperformed methods based on traditional deep features. The introduction of the proposed framework has improved the performance in AlexNet almost by a margin of 14\%. Proposed method based on VGG produces the best performance on this dataset by introducing a 4.5\% of an improvement as compared with the Adjusted Graphical Method proposed in \cite{Saleh:2013:OAD:2514950.2516141}. 

\begin{table}[htp!]
	\centering
		\caption{Abnormal image detection results on the 1001 Abnormal Objects dataset.}
	\label{table:ab1001}
	\resizebox{.9\linewidth}{!}{
		\begin{tabular} {|l|c|c|}
			\hline
			Method & AUC (Std. Dev.)  \\ \hline\hline
			Graphical Model \cite{Saleh:2013:OAD:2514950.2516141} & 0.870  \\
			Adjusted Graphical Model \cite{Saleh:2013:OAD:2514950.2516141} & 0.911\\
			Autoencoder\cite{MNISTAUTO} & 0.674 (0.120)  \\
			OCNN AlexNet\cite{ocnn} &  0.845 (0.148) \\
			OCNN VGG16\cite{ocnn} &  0.888 (0.0460)\\ 
			\hline
			AlexNet Features KNN& 0.790 (0.074) \\
			VGG16 Features KNN& 0.847  (0.074)  \\
			AlexNet Binary KNN & 0.621  (0.153) \\  
			VGG16 Binary KNN& 0.848  (0.081)  \\
			\hline
			AlexNet Features IF& 0.613 (0.085) \\
			VGG16 Features IF& 0.731  (0.078)  \\
			AlexNet Binary IF & 0.641  (0.089)\\  
			VGG16 Binary IF& 0.715  (0.077)  \\ \hline
			
			AlexNet Features SVM& 0.732 (0.094) \\
			VGG16 Features SVM& 0.847  (0.074) \\
			AlexNet Binary SVM & 0.736  (0.115)\\  
			VGG16 Binary SVM& 0.834  (0.083) \\\hline
			
			AlexNet Features GMM& 0.679 (0.103) \\
			VGG16 Features GMM& 0.818  (0.072)  \\
			AlexNet Binary GMM & 0.696  (0.116)\\  
			VGG16 Binary GMM& 0.803  (0.103) \\\hline
			
			DOC AlexNet (ours) & 0.930 (0.032) \\
			DOC VGG16 (ours) & \textbf{0.956 (0.031)}  \\
			\hline

		\end{tabular} 
	}
\end{table}

\noindent {\bf{One-Class Novelty Detection:}}
In one-class novelty detection, the goal is to assess the novelty of a new sample based on previously observed samples. Since novel examples do not exist prior to test time,  training is carried out using one-class learning principles. In the previous works \cite{Bodesheim_2013_CVPR},\cite{DBLP:conf/ijcnn/2016}, the performance of novelty detection has been assessed based on different classes of the ImageNet and the Caltech 256 datasets. Since all CNNs used in our work have been trained using the ImageNet dataset, we use the Caltech 256 dataset to evaluate the performance of one-class novelty detection.  The Caltech 256 dataset contains images belonging to 256 classes with total of 30607 images. In our experiments, each single class was considered separately and all other classes were considered as alien.  Sample images belonging to three classes in the dataset are shown in Figure~\ref{fig:samples} (b). First, consistent with the protocol described in \cite{DBLP:conf/ijcnn/2016}, AUC of 20 random repetitions were evaluated by considering the \textit{American Flag} class as the known class and  by considering boom-box, bulldozer and can-non classes as alien classes.  Results corresponding to different methods are  tabulated in Table~\ref{table:novflag}. 

In order to evaluate the robustness of our method, we carried out an additional test involving all classes of the Caltech 256 dataset. In this test, first a single class is chosen to be the enrolled class. Then, the effectiveness of the learned classifier was evaluated by considering samples from all other 255 classes. We did 40 iterations of the same experiment by considering first 40 classes of the Caltech 256 dataset one at a time as the enrolled class. Since there are 255 alien classes in this test as opposed to the first test, where there were only three alien classes, performance is expected to be lower than in the former.  Results of this experiment are tabulated in Table~\ref{table:nov}.

\begin{table}[htp!]
	\centering
		\caption{Novelty detection results on the Caltech 256 where \textit{American Flag} class is taken as the known class.}
	\label{table:novflag}
	\resizebox{0.8\linewidth}{!}{
		\begin{tabular} {|l|c|}
			\hline
			Method & AUC (Std. Dev.)\\ \hline \hline
			One Class SVM  \cite{Scholkopf:2001:ESH:1119748.1119749} &0.606 (0.003) \\
			KNFST \cite{Bodesheim_2013_CVPR} & 0.575 (0.004)  \\
			Oc-KNFD \cite{DBLP:conf/ijcnn/2016} & 0.619 (0.003) \\
			Autoencoder\cite{MNISTAUTO} & 0.532(0.003)  \\
					
			OCNN AlexNet\cite{ocnn} & 0.907 (0.029) \\
			OCNN VGG16\cite{ocnn} &  0.943 (0.035)\\ 
			
			\hline
			AlexNet Features KNN& 0.811 (0.003)  \\
			VGG16 Features KNN& 0.951 (0.023) \\
			
			AlexNet Binary KNN& 0.920 (0.026)  \\
			VGG16 Binary KNN& 0.997 (0.001) \\

			\hline
		
			AlexNet Features IF& 0.836 (0.005) \\
			VGG16 Features IF& 0.910  (0.035)  \\
			AlexNet Binary IF & 0.795  (0.007)\\  
			VGG16 Binary IF& 0.907  (0.033)  \\ \hline
			
			AlexNet Features SVM& 0.878 (0.007) \\
			VGG16 Features SVM& 0.951  (0.029) \\
			AlexNet Binary SVM & 0.920  (0.008)\\  
			VGG16 Binary SVM& 0.942  (0.031) \\\hline
			
			AlexNet Features GMM& 0.842 (0.004) \\
			VGG16 Features GMM& 0.901  (0.023)  \\
			AlexNet Binary GMM & 0.860  (0.009)\\  
			VGG16 Binary GMM& 0.924  (0.025) \\\hline

		DOC AlexNet (ours) & 0.930 (0.005) \\
		DOC VGG16 (ours) & \textbf{0.999 (0.001)} \\
			\hline
			
		\end{tabular} 
	}
\end{table}

It is evident from the results in Table~\ref{table:novflag} that a significant improvement is obtained in the proposed method compared to previously proposed methods. However, as shown in Table~\ref{table:nov} this performance is not limited just to a American Flag. Approximately the same level of performance is seen across all classes in the Caltech 256 dataset.  Proposed method has improved the performance of AlexNet by nearly 13\% where as the improvement the proposed method has had on VGG16 is around 9\%. It is interesting to note that binary CNN classifier based on the VGG framework has recorded performance very close to the proposed method in this task (difference in performance is about 1\%). This is due to the fact that both ImageNet and Caltech 256 databases contain similar object classes. Therefore, in this particular case, ImageNet samples are a good representative of novel object classes present in Caltech 256. As a result of this special situation, binary CNN is able to produce results on par with the proposed method. However, this result does not hold true in general as evident from other two experiments.

\begin{table}[htp!]
	\centering
	\caption{Average Novelty detection results on the Caltech 256 dataset.}
	\label{table:nov}
	\resizebox{0.8\linewidth}{!}{
	
		\begin{tabular} {|l|c|c|}
			\hline 
			Method &AUC (Std. Dev.)\\ \hline \hline
			One Class SVM  \cite{Scholkopf:2001:ESH:1119748.1119749} & 0.531 (0.120)\\
			Autoencoder\cite{MNISTAUTO} & 0.623 (0.072)\\
					OCNN AlexNet\cite{ocnn} & 0.826 (0.153) \\
			OCNN VGG16\cite{ocnn} &  0.885 (0.144)\\ 
			\hline
			AlexNet Features KNN& 0.820 (0.062)\\
			VGG16 Features KNN& 0.897 (0.050) \\
			AlexNet Binary KNN& 0.860 (0.065) \\
			VGG16 Binary KNN& 0.902 (0.024) \\ 
			
					\hline
			
			AlexNet Features IF& 0.794 (0.075) \\
			VGG16 Features IF& 0.890 (0.049)  \\
			AlexNet Binary IF & 0.788 (0.087)\\  
			VGG16 Binary IF& 0.891  (0.053)  \\ \hline
			
			AlexNet Features SVM& 0.852 (0.057) \\
			VGG16 Features SVM& 0.902 (0.050) \\
			AlexNet Binary SVM & 0.856  (0.058)\\  
			VGG16 Binary SVM& 0.909  (0.047) \\\hline
			
			AlexNet Features GMM& 0.790 (0.083) \\
			VGG16 Features GMM& 0.852  (0.087)  \\
			AlexNet Binary GMM & 0.801  (0.083)\\  
			VGG16 Binary GMM& 0.870  (0.069) \\\hline

			DOC AlexNet (ours) & 0.959 (0.021) \\
			DOC VGG16 (ours) & \textbf{0.981 (0.022)} \\
			\hline
			
		\end{tabular} 
	}

\end{table}

\noindent {\bf{Active Authentication (AA):}} In the final set of tests, we evaluate the performance of different methods on the UMDAA-02 mobile AA dataset \cite{UMDAA02}.  
The UMDAA-02 dataset contains multi-modal sensor observations captured over a span of two months from 48 users for the problem of continuous authentication. In this experiment, we only use the face images of users collected by the front-facing camera of the mobile device.  The UMDAA-02 dataset is a highly challenging dataset with large amount of intra-class variation including pose, illumination and appearance variations.  Sample images from the UMDAA-02 dataset are shown in Figure~\ref{fig:samples} (c).  As a result of these high degrees of variations, in some cases the inter-class distance between different classes seem to be comparatively lower making recognition challenging.

During testing, we considered first 13 users taking one user at a time to represent the enrolled class where all the other users were considered to be alien.  The performance of different methods on this dataset is tabulated in Table~\ref{table:AA}. 

\begin{table}[htp!]
	\centering
	\caption{Active Authentication results on the UMDAA-02 dataset.}
	\label{table:AA}
	\resizebox{0.8\linewidth}{!}{
	
		\begin{tabular} {|l|c|c|}
			\hline
			Method & AUC (Std. Dev.)\\ \hline \hline
			One Class SVM  \cite{Scholkopf:2001:ESH:1119748.1119749} & 0.594 (0.070) \\
			Autoencoder\cite{MNISTAUTO}  & 0.643 (0.074)\\
					OCNN AlexNet\cite{ocnn} & 0.595 (0.045) \\
			OCNN VGG16\cite{ocnn} &  0.574 (0.039)\\ 
			\hline
			AlexNet Features KNN& 0.708 (0.060) \\
			VGG16 Features KNN& 0.748 (0.082)  \\
			AlexNet Binary KNN& 0.627  (0.128)  \\
			VGG16 Binary KNN & 0.687 (0.086) \\
			
			\hline
			
				AlexNet Features IF& 0.694 (0.075) \\
			VGG16 Features IF& 0.733 (0.080)  \\
			AlexNet Binary IF & 0.625 (0.099)\\  
			VGG16 Binary IF& 0.677  (0.078)  \\
			\hline
			
			AlexNet Features SVM& 0.702 (0.087) \\
			VGG16 Features SVM& 0.751 (0.075) \\
			AlexNet Binary SVM & 0.656  (0.112)\\  
			VGG16 Binary SVM& 0.685  (0.076) \\\hline
			
			AlexNet Features GMM& 0.690 (0.077) \\
			VGG16 Features GMM& 0.751  (0.082)  \\
			AlexNet Binary GMM & 0.629  (0.110)\\  
			VGG16 Binary GMM& 0.650 (0.087) \\\hline
			
			DOC AlexNet (ours) & 0.786 (0.061)  \\
			DOC VGG16 (ours) & \textbf{0.810 (0.067)}\\
			
			\hline
			
		\end{tabular} 
	}
	
\end{table}

Recognition results are comparatively lower for this task compared to the other tasks considered in this paper. This is both due to the nature of the application and the dataset. However, similar to the other cases, there is a significant performance improvement in proposed method compared to the conventional CNN-based methods.  In the case of AlexNet, improvement induced by the proposed method is nearly 8\% whereas it is around 6\% for VGG16.  The best performance is obtained by the proposed method based on the VGG16 network.

\subsection{Discussion}

\begin{figure}[htp!]
	\centering
	\includegraphics[width=\linewidth]{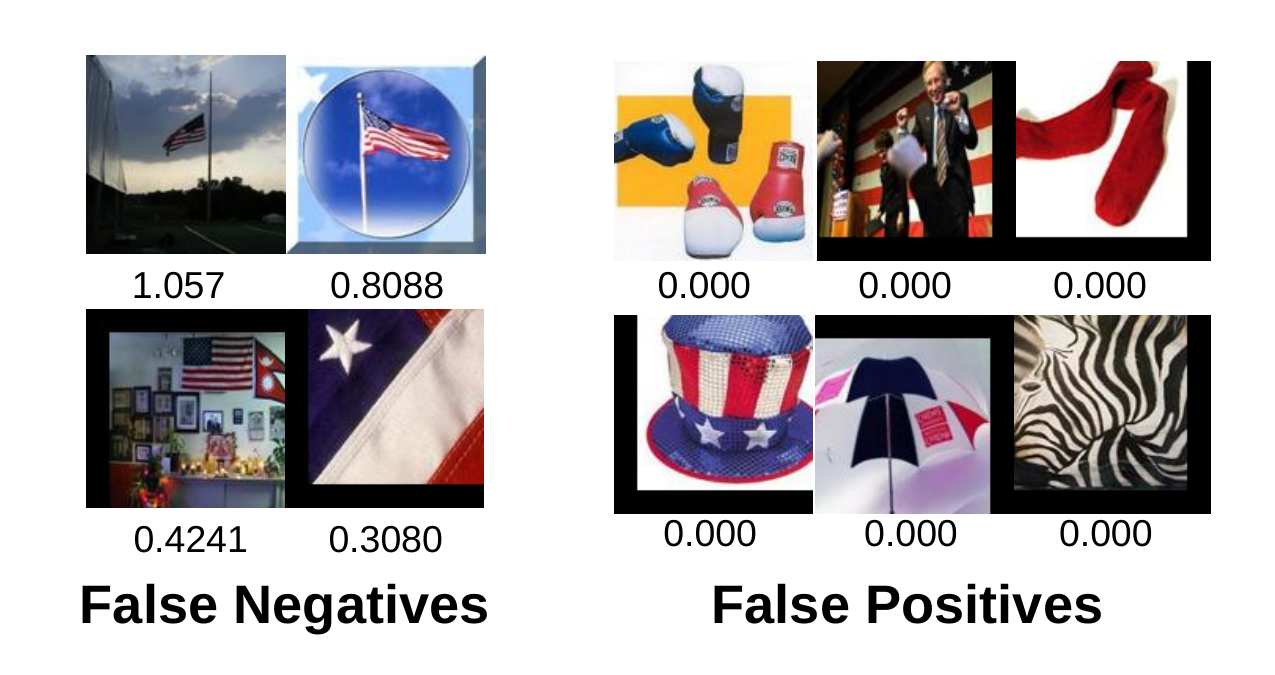}\hskip30pt
	\vskip -8pt\caption{Sample false detections for the one-class problem of novelty detection where the class of interest is \textit {American Flag} (Detection threshold $\delta$ = 0). Obtained Euclidean distance scores are also shown in the figure.}
	\label{fig:failed_cases}
\end{figure}

\noindent {\bf{Analysis on mis-classifications:}}
The proposed method produces better separation between the class under consideration and alien samples as presented in the results section. However, it is interesting to investigate on what conditions the proposed method fails. Shown in Figure~\ref{fig:failed_cases} are a few cases where the proposed method produced erroneous detections for the problem of one-class novelty detection with respect to the \textit{American Flag} class (in this experiment, all other classes in Caltech256 dataset were used as alien classes). Here, detection threshold has been selected as $\delta = 0$. Mean detection scores for \textit{American Flag} and alien images were 0.0398 and 8.8884, respectively. 

As can be see from Figure~\ref{fig:failed_cases}, in majority of false negative cases, the American Flag either appears in the background of the image or it is too closer to clearly identify its characteristics. On the other hand, false positive images either predominantly have American flag colors or the texture of a waving flag. It should be noted that the nature of mis-classifications obtained in this experiment are very similar to that of multi-class CNN-based classification.\\

\noindent {\bf{Using a subset of the reference dataset:}}
In practice, the reference dataset is often enormous in size. For example, the ImageNet dataset has in excess of one million images. Therefore, using the whole reference dataset for transfer learning may be inconvenient. Due to the low number of training iterations required, it is possible to use a subset of the original reference dataset in place of the reference dataset without causing over-fitting. In our experiments, training of the reference network was done using the validation set of the ImageNet dataset. Recall that initially, both networks are loaded with pre-trained models. It should be noted that these pre-trained models have to be trained using the whole reference dataset. Otherwise, the resulting network will have poor generalization properties.\\

\noindent {\bf{Number of training iterations:}}
In an event when only a subset of the original reference dataset is used, the training process should be closely monitored. It is best if training can be terminated as soon as the  composite loss converges. Training the network long after composite loss has converged could result in inferior features due to over-fitting.  This is the trade-off of using a subset of the reference dataset.  In our experiments, convergence occurred around 2 epochs for all test cases (Figure~\ref{fig:output_trainingLoss}). We used a fixed number of iterations (700) for each dataset in our experiments.\\ 

\noindent {\bf{Effect of number of templates:}}
In all conducted experiments, we fixed the number of templates used for recognition to 40. In order to analyze the effect of template size on the performance of our method, we conducted an experiment by varying the template size. We considered two cases: first, the novelty detection problem related to the \textit{American Flag} (all other classes in Caltech256 dataset were used as alien classes), where the recognition rates were very high at 98\%; secondly, the AA problem where the recognition results were modest. We considered Ph01USER002 from the UMDAA-02 dataset for the study on AA. We carried out twenty iterations of testing for each case. The obtained results are tabulated in Table~\ref{table:sizematters}.

\begin{table*}[htp!]
	\centering
	\caption{Mean AUC (with standard deviation values in brackets) obtained for different template sizes. }
	\label{table:sizematters}
	\resizebox{0.8\linewidth}{!}{
		\begin{tabular} {|l|c|c|c|c|c|c|}
			\hline
			Number of templates & 1 & 5 & 10 & 20 & 30 & 40\\ \hline
			American Flag &  0.987 &  \textbf{0.988} & 0.987 & 0.988 & 0.988 & 0.988 \\
			(Caltech 256) &  (0.0034)
			& (0.0032)
			& (0.0030)
			& (0.0038)
			& (0.0029)
			& (0.0045) \\
			
			\hline
			Ph01USER002  & 0.762 &  0.788 & 0.806 & 0.787 & 0.821 & \textbf{0.823}  \\
			(UMDAA02) &  (0.0226)
			& (0.0270)
			& (0.0134)
			& (0.0262)
			& (0.0165)
			& (0.0168)\\
			
			\hline
			
		\end{tabular} 
	}
	
\end{table*}

According to the results in Table~\ref{table:sizematters},  it appears that when the proposed method is able to isolate a class sufficiently, as in the case of novelty detection, the choice of the number of templates is not important. Note that even a single template can generate significantly accurate results. However, this is not the case for AA. Reported relatively lower AUC values in testing suggests that all faces of different users lie in a smaller subspace. In such a scenario, using more templates have generated better AUC values.

\subsection{Impact of Different Features}
In this subsection, we investigate the impact of different choices of $h_c$ and $g$ has on the recognition performance. Feature was varied from $fc6$ to $fc8$ and the performance of the abnormality detection task was evaluated. When $fc6$ was used as the feature, the sub-network $g$ consisted layers from $\text{conv}1$ to $fc6$, where layers $fc7$ and $fc8$ were associated with the sub network  $h_c$. Similarly, when the layer $fc7$ was considered as the feature, the sub-networks $g$ and $h_c$ consisted of layers $\text{conv}1-fc7$ and $fc8$, respectively. 

In Table~\ref{table:ablationf}, the recognition performance on abnormality image detection task is tabulated for different choices of $h_c$ and $g$. From Table~\ref{table:ablationf} we see that in both AlexNet and VGG16 architectures, extracting features at a later layer has yielded in better performance in general. For example, for VGG16 extracting features from $fc6, fc7$ and $fc8$ layers has yielded AUC of $0.856, 0.956$ and $0.969$, respectively. This observation is not surprising on two accounts. First, it is well-known that later layers of deep networks result in better generalization. Secondly, \textit{Compactness Loss} is minimized with respect to features of the target dataset extracted in the $fc8$ layer. Therefore, it is expected that $fc8$ layer provides better compactness in the target dataset.
\begin{table*}[htp!]
	\centering
	\caption{Abnormal image detection results for different choices of the reference dataset.}
	\label{table:ablationf}
	\resizebox{.7\linewidth}{!}{
		\begin{tabular} {|l|c|c|c|c|c|c|}
			\hline
			& fc6 & fc7 & fc8 \\ \hline\hline
				
			DOC AlexNet (ours) & 0.936 (0.041) & 0.930 (0.032) &  \textbf{0.947 (0.035)}
			\\
			DOC VGG16 (ours)& 0.856 (0.118)  & 0.956 (0.031) &  \textbf{0.969 (0.029)}
			\\
			\hline		
		\end{tabular} 
	}
\end{table*}

\subsection{Impact of the Reference Dataset}
The proposed method utilizes a reference dataset to ensure that the learned feature is informative by minimizing the \textit{descriptiveness loss}. For this scheme to result in effective features, the reference dataset has to be a non-trivial multi-class object dataset.  In this subsection, we investigate the impact of the reference dataset on the recognition performance. In particular, abnormal image detection experiment on  the Abnormal 1001 dataset was repeated with a different choice of the reference dataset. In this experiment ILSVRC12 \cite{imagenet_cvpr09}, Places365 \cite{zhou2017places} and Oxford Flowers 102 \cite{Nilsback08} datasets were used as the reference dataset. We used publicly available pre-trained networks from caffe model zoo \cite{caffe} in our evaluations.

In Table~\ref{table:ablationref} the recognition performance for the proposed method as well as the baseline methods are tabulated for each considered dataset. From Table~\ref{table:ablationref} we observe that the recognition performance has dropped when a different reference dataset is used in the case of VGG16 architecture. Places365 has resulted in a drop of 0.038 whereas the Oxford flowers 102 dataset has resulted in a drop of 0.026. When the AlexNet architecture is used, a similar trend can be observed. Since Places365 has smaller number of classes than ILVRC12, it is reasonable to assume that the latter is more diverse in content. As a result, it has helped the network to learn more informative features. On the other hand, although Oxford flowers 102 has even fewer classes, it should be noted that it is a fine-grain classification dataset. As a result, it too has helped to learn more informative features compared to Places365. However, due to the presence of large number of non-trivial classes, the ILVRC12 dataset has yielded the best performance among the considered cases.

\begin{table*}[htp!]
	\centering
	\caption{Abnormal image detection results for different choices of the reference dataset.}
	\label{table:ablationref}
	\resizebox{.7\linewidth}{!}{
		\begin{tabular} {|l|c|c|c|c|c|c|}
			\hline
			  & ILVRC12& Places365 & Flowers102 \\ \hline\hline
			
			AlexNet Features KNN& 0.790 (0.074) & 0.856	(0.056) &	0.819	(0.075)
			\\
			VGG16 Features KNN& 0.847  (0.074) & 0.809	(0.100) &	0.828	(0.077)
			\\
			AlexNet Binary KNN & 0.621  (0.153) & 0.851	(0.060) &	0.823	(0.084)
			\\  
			VGG16 Binary KNN& 0.848  (0.081)  & 0.837	(0.090)	 & 0.839	(0.077)
			\\
			\hline
			AlexNet Features IF& 0.613 (0.085) & 0.771	(0.107)	& 0.739	(0.098)
			\\
			VGG16 Features IF& 0.731  (0.078) & 0.595	(0.179) &	0.685	(0.154)
			\\
			AlexNet Binary IF & 0.641  (0.089) & 0.777	(0.092)	& 0.699	(0.129)
			\\  
			VGG16 Binary IF& 0.715  (0.077) & 0.637	(0.159)	& 0.777	(0.110)
			\\ \hline
			
			AlexNet Features SVM& 0.732 (0.094) & 0.839	(0.062)	& 0.818	(0.076)
			\\
			VGG16 Features SVM& 0.847  (0.074) & 0.776	(0.113) &	0.826	(0.077)
			\\
			AlexNet Binary SVM & 0.736  (0.115) & 0.847	(0.065) &	0.823	(0.083)
			\\  
			VGG16 Binary SVM& 0.834  (0.083) & 0.789	(0.114)	& 0.788	(0.089)
			\\\hline
			
			AlexNet Features GMM& 0.679 (0.103) & 0.832	(0.069) &	0.779	(0.076)
			\\
			VGG16 Features GMM& 0.818  (0.072) & 0.782	(0.103)	& 0.771	(0.114)
			\\
			AlexNet Binary GMM & 0.696  (0.116) & 0.835	(0.068)	& 0.815	(0.101)
			\\  
			VGG16 Binary GMM & 0.803  (0.103) & 0.770	(0.103)	& 0.777	(0.110)
			\\\hline
			
			DOC AlexNet (ours) & 0.930 (0.032) &  0.896	(0.019 )&	0.899	(0.052)
			\\
			DOC VGG16 (ours) & \textbf{0.956 (0.031)} & \textbf{0.918	(0.049)}	&  \textbf{0.930	(0.059)}
			\\
			\hline

		\end{tabular} 
	}
\end{table*}

\section{Conclusion}\label{sec:con}
We introduced a deep learning solution for the problem of one-class classification, where only training samples of a single class are available during training.  We proposed a feature learning scheme that engineers class-specific features that are generically discriminative. To facilitate the learning process, we proposed two loss functions \textit{descriptiveness loss} and \textit{compactness loss} with a CNN network structure. Proposed network structure could be based on any CNN backbone of choice. The effectiveness of the proposed method is shown in results for AlexNet and VGG16-based backbone architectures. The performance of the proposed method is tested on publicly available datasets for abnormal image detection, novelty detection and face-based mobile active authentication. The proposed method obtained the state-of-the-art performance in each test case.

\section*{Acknowledgement}
This  work  was  supported  by US Office of Naval Research (ONR) Grant YIP
N00014-16-1-3134. 

\section*{Appendix A: Derivations }

\noindent {\bf{Batch-variance Loss is a Scaled Version of Sample Variance:}} 
Consider the definition of batch-variance loss defined as,

$ l_b = \frac{1}{nk}  \sum_{i=1}^{n} \mathbf{z_i}^T\mathbf{z_i}$ where, $ \mathbf{z_i}  = \bigg[ \mathbf{x_i} - \frac{1}{n-1}\sum_{j \neq i}\mathbf{x_j}\bigg]$.
Re-arranging terms in $\mathbf{z_i}$,
$$ \mathbf{z_i}  = \bigg[ \mathbf{x_i} - \frac{1}{n-1}\sum_{j=1}^{n}\mathbf{x_j} + \frac{1}{n-1} \mathbf{x_i} \bigg]$$

$$ \mathbf{z_i}  = \bigg[\frac{n}{n-1} \mathbf{x_i} - \frac{1}{n-1}\sum_{j=1}^{n}\mathbf{x_j}  \bigg]$$

$$ \mathbf{z_i}  = \frac{n}{n-1} \bigg[ \mathbf{x_i} - \frac{1}{n}\sum_{j=1}^{n}\mathbf{x_j}  \bigg]$$

$$ \mathbf{{z_i}}^T{\mathbf{z_i}}  = \frac{n^2}{(n-1)^2} \bigg[ \mathbf{x_i} - \frac{1}{n}\sum_{j=1}^{n}\mathbf{x_j}  \bigg]^T \bigg[ \mathbf{x_i} - \frac{1}{n}\sum_{j=1}^{n}\mathbf{x_j}  \bigg]$$
But, $\bigg[ \mathbf{x_i} - \frac{1}{n}\sum_{j=1}^{n}\mathbf{x_j}  \bigg]^T \bigg[ \mathbf{x_i} - \frac{1}{n}\sum_{j=1}^{n}\mathbf{x_j}  \bigg]$ is the sample variance $\sigma_i^2$. Therefore,

$$l_b  = \frac{1}{nk}  \sum_{i=1}^{n} \frac{n^2 \sigma_i^2}{(n-1)^2}  $$

Therefore, $l_b  = \beta \sigma^2 $, where $\beta = \frac{n^2}{k(n-1)^2}$ is a constant and $\sigma^2 $ is the average sample variance.

\noindent {\bf{Backpropagation of Batch-variance Loss:}} \label{proof1}

Consider the definition of batch variance loss $l_b$,

$ l_b = \frac{1}{nk}  \sum_{i=1}^{n} \mathbf{z_i}^T\mathbf{z_i}$ where, $ \mathbf{z_i}  = \mathbf{x_i} - \mathbf{m_i}.$

From the definition of the inner product,

$\mathbf{{z_i}^T}\mathbf{{z_i}} = \sum_{j=1}^{k} {z_{ij}}^2.$ Therefore, $l_b$ can be written as,

$$l_b = \frac{1}{nk} \sum_{i=1}^{n} \sum_{l=1}^{k} (x_{il}-m_{il})^2.$$

Taking partial derivatives of $l_b$ with respect to ${x_{ij}}$. By chain rule we obtain,
$$\frac{\partial l_b}{\partial {x_{ij}}} = \frac{2}{nk} \sum_{l=1}^{k} x_{il}-m_{il}  \times \frac{\partial x_{il}-m_{il}}{\partial {x_{ij}}}. $$

Note that $\frac{\partial x_{ij}-m_{ij}}{\partial {x_{ij}}} = 1$ when $j=l$. Otherwise, $\frac{\partial x_{ij}-m_{ij}}{\partial {x_{ij}}} = -\frac{\partial m_{ij}}{\partial {x_{ij}}} = \frac{-1}{n-1}.$

$$\frac{\partial l_b}{\partial {x_{ij}}} = \frac{2}{nk} \bigg[ x_{ij}-m_{ij} - \frac{1}{n-1}\sum_{l \neq j} x_{il}-m_{il}   \bigg].$$

$$\frac{\partial l_b}{\partial {x_{ij}}} = \frac{2}{nk} \bigg[\frac{n}{n-1} x_{ij}-m_{ij} - \frac{1}{n-1}\sum_{l=1}^{n} x_{il}-m_{il}   \bigg].$$

$$\frac{\partial l_b}{\partial {x_{ij}}} = \frac{2}{(n-1)nk} \bigg[n \times (x_{ij}-m_{ij}) - \sum_{l=1}^{n} (x_{il}-m_{il} )  \bigg].$$

\section*{Appendix B : Detailed Network Architectures }
\label{detailedarchi}
\begin{figure}[htp!]
	\centering
	\includegraphics[width=1\linewidth]{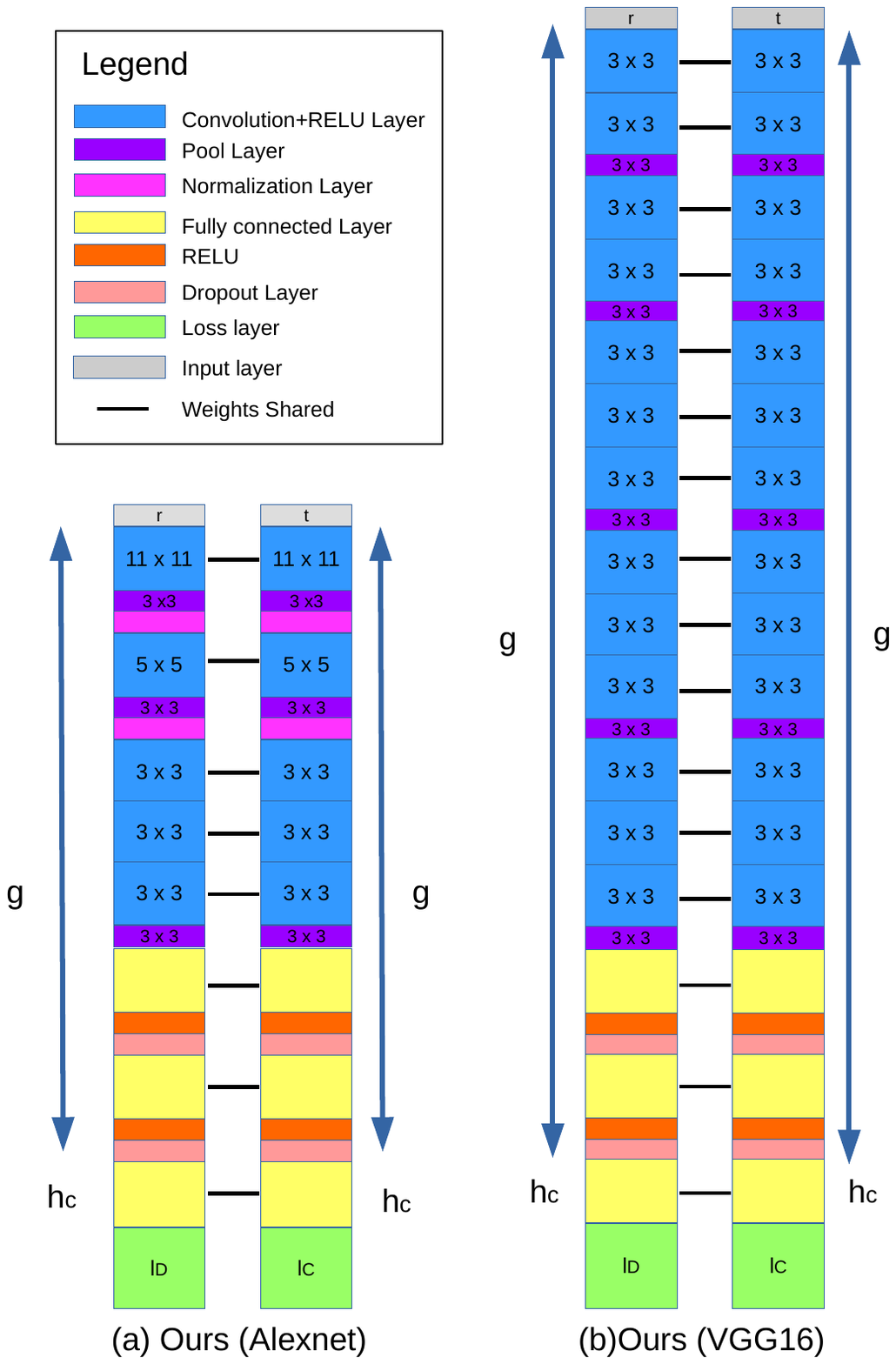}\hskip30pt
	\vskip -0pt\caption{CNN architectures based on AlexNet and VGG16 backbones for the proposed method.}
	\label{fig:sum}
\end{figure}

Shown in Figure~\ref{fig:sum} are the adaptations of the proposed method to the existing CNN architectures- (a) AlexNet and (b) VGG16, respectively. We have used these two architectures to carry out all experiments in this paper. In both architectures, two images from the target dataset $t$ and the reference dataset $r$ are fed into the network to evaluate cross entropy loss and batch-variance loss. Weights of convolution and fully-connected layers are shared between the two branches of the network. For all experiments, stochastic gradient descent algorithm is used with a learning rate of $5 \times 10^{-5}$ and a weight decay of 0.0005.

{\small
	\bibliographystyle{ieee}
	\bibliography{egbib}
}

\begin{IEEEbiography}[{\includegraphics[width=1in,height=1.25in,clip,keepaspectratio]{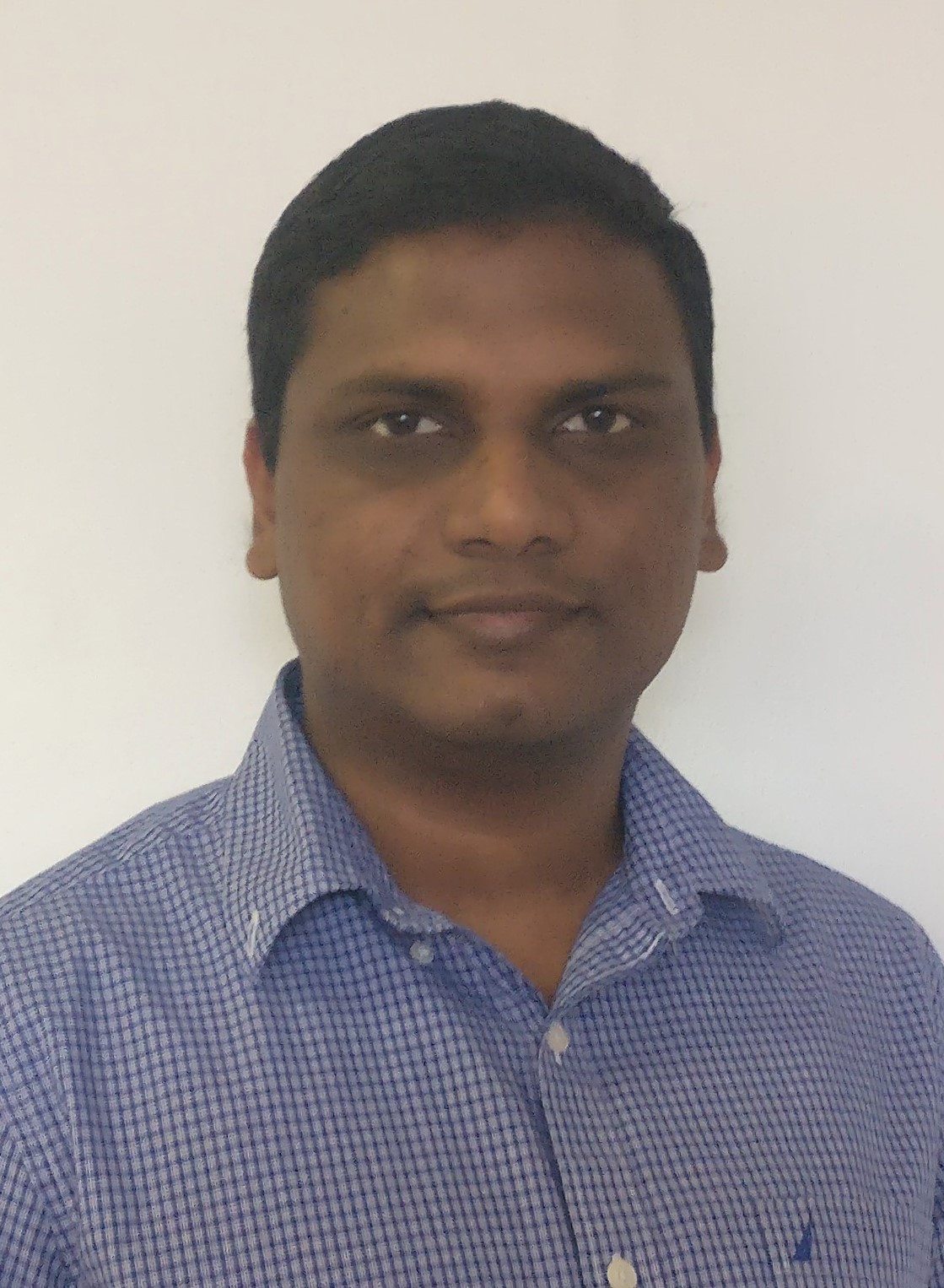}}]{Pramuditha Perera}
	Pramuditha Perera is a Ph.D. student at the Department of Electrical and Computer Engineering, Johns Hopkins University, Baltimore, USA. He received his bachelors degree in Electrical and Electronic Engineering from University of Peradeniya, Sri Lanka in 2014. He completed his masters degree in Electrical and Computer Engineering at Rutgers University, USA in 2018. His research interests include computer vision and machine learning with applications in biometrics. His work received the best student paper award at IAPR ICPR 2018.  
\end{IEEEbiography}

\begin{IEEEbiography}[{\includegraphics[width=1in,height=1.25in,clip,keepaspectratio]{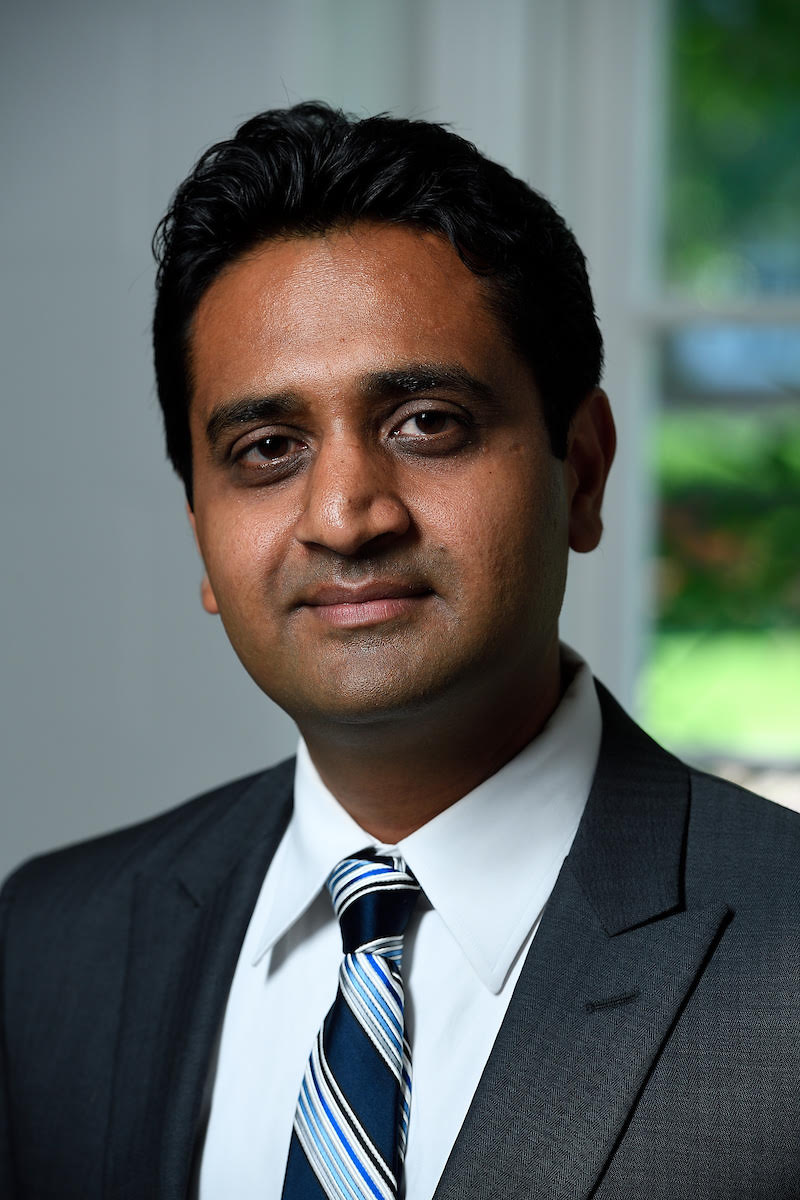}}]{Vishal M. Patel}
	Vishal M. Patel [SM'15] is an Assistant Professor in the Department of Electrical and Computer Engineering (ECE) at Johns Hopkins University.  Prior to joining Hopkins, he was an A. Walter Tyson Assistant Professor in the Department of ECE at Rutgers University and a member of the research faculty at the University of Maryland Institute for Advanced Computer Studies (UMIACS). He completed his Ph.D. in Electrical Engineering from the University of Maryland, College Park, MD, in 2010. His current research interests include signal processing, computer vision, and pattern recognition with applications in biometrics and imaging. He has received a number of awards including the 2016 ONR Young Investigator Award, the 2016 Jimmy Lin Award for Invention, A. Walter Tyson Assistant Professorship Award, Best Paper Award at IEEE AVSS 2017, Best Paper Award at IEEE BTAS 2015, Honorable Mention Paper Award at IAPR ICB 2018, two Best Student Paper Awards at IAPR ICPR 2018, and Best Poster Awards at BTAS 2015 and 2016. He is an Associate Editor of the IEEE Signal Processing Magazine, IEEE Biometrics Compendium, and serves on the Information Forensics and Security Technical Committee of the IEEE Signal Processing Society. He is a member of Eta Kappa Nu, Pi Mu Epsilon, and Phi Beta Kappa. 
\end{IEEEbiography}
\end{document}